\documentclass[letterpaper, 10 pt, conference]{ieeeconf}  

\IEEEoverridecommandlockouts                              

\usepackage[acronym]{glossaries}
\usepackage{multirow}
\usepackage{booktabs} 
\usepackage[colorlinks=true]{hyperref} 
\usepackage{amsfonts}
\usepackage{soul}
\usepackage{diagbox}
\usepackage[hang,flushmargin,symbol]{footmisc}
\usepackage{adjustbox}
\usepackage{microtype}
\usepackage{array}
\usepackage{cite}
\usepackage{pifont}
\let\labelindent\relax
\usepackage{enumitem}

\newcolumntype{R}[2]{%
    >{\adjustbox{angle=#1,lap=\width-(#2)}\bgroup}%
    l%
    <{\egroup}%
}
\newcommand*\rot{\multicolumn{1}{R{45}{1em}}}
\newcommand*\multitask{$^\text{{\tiny MTL}}$}

\usepackage[capitalize]{cleveref}
\crefname{section}{Sec.}{Secs.}
\Crefname{section}{Section}{Sections}
\Crefname{table}{Table}{Tables}
\crefname{table}{Tab.}{Tabs.}
\crefname{algorithm}{Algo.}{Algos.}

\newcommand{\figref}[1]{Fig.~\ref{#1}}
\newcommand{\tabref}[1]{Tab.~\ref{#1}}
\newcommand{\secref}[1]{Sec.~\ref{#1}}

\def\ie{\emph{i.e}.}

\usepackage[dvipsnames]{xcolor}
\definecolor{barrier}{RGB}{255, 240, 150}
\definecolor{bicycle}{RGB}{100, 230, 245}
\definecolor{bus}{RGB}{100, 80, 250}
\definecolor{car}{RGB}{100, 150, 245}
\definecolor{construction-vehicle}{RGB}{80, 20, 160}
\definecolor{motorcycle}{RGB}{30, 60, 150}
\definecolor{pedestrian}{RGB}{155, 30, 30}
\definecolor{traffic-cone}{RGB}{255, 94, 0}
\definecolor{trailer}{RGB}{63, 81, 181}
\definecolor{truck}{RGB}{80, 30, 180}
\definecolor{driveable-surface}{RGB}{255, 0, 255}
\definecolor{other-flat}{RGB}{175, 0, 75}
\definecolor{sidewalk}{RGB}{75, 0, 75}
\definecolor{terrain}{RGB}{120, 200, 40}
\definecolor{manmade}{RGB}{255, 200, 0}
\definecolor{vegetation}{RGB}{0, 140, 0}

\newacronym{rpn}{RPN}{Region Proposal Network}
\newacronym{roi}{RoI}{Region of Interest}
\newacronym{iou}{IoU}{intersection-over-union}
\newacronym{nms}{NMS}{non-maximum suppression}
\newacronym{fpn}{FPN}{Feature Proposal Network}
\newacronym{fcn}{FCN}{fully-connected network}
\newacronym{rcgn}{RCGN}{Relational Graph Neural Network}
\newacronym{ggnn}{GGNN}{Gated-Graph Neural Network}
\newacronym{mlp}{MLP}{Multi-Layer Perceptron}
\newacronym{svg}{SVG}{singular value decomposition}
\newacronym{kge}{KGE}{knowledge graph embedded}
\newacronym{sota}{SoTA}{state-of-the-art}
\newacronym{mlr}{MLR}{multi logistic regression}
\newacronym{vos}{VOS}{visualizing similarities between objects}
\newacronym{mot}{MOT}{Multi-Object Tracking}
\newacronym{ssmot}{SSMOT}{Self-Supervised Multi-Object Tracking}
\newacronym{mot17}{MOT17}{~\cite{MOT16}}
\newacronym{ssl}{SSL}{self-supervised learning}
\newacronym{fps}{FPS}{frames per second}
\newacronym{msa}{MSA}{minimum-sum assigment}
\newacronym{tempo}{TempO}{temporal ordering pretext}
\newacronym{nn}{NN}{nearest-neighbor}
\newacronym{reid}{ReID}{re-identification}
\newacronym{vit}{ViT}{vision transformer} 
\newacronym{subco}{SubCo}{subsample consistency}
\newacronym{rnn}{RNN}{recurrent neural network}
\newacronym{lidar}{LiDAR}{LiDAR}
\newacronym{semkitti}{SemanticKITTI}{SemanticKITTI~\cite{semantickitti}}
\newacronym{nuscenes}{nuScenes}{nuScenes~\cite{nuscenes}}
\newacronym{bev}{BEV}{bird's-eye-view}
\newacronym{kitti}{KITTI}{KITTI~\cite{kitti}}
\newacronym{patt}{PAtt}{point-attention}
\newacronym{fov}{FOV}{field-of-view}
\newacronym{kpconv}{KPConv}{~\cite{kpconv}}
\newacronym{map}{mAP}{mean average precision}
\newacronym{aos}{AOS}{average orientation similarity}
\newacronym{semantickitti}{SemanticKITTI}{SemanticKITTI~\cite{semantickitti}}
\newacronym{knn}{kNN}{k-nearest neighbors}
\newacronym{tarl}{TARL}{TARL~\cite{tarl}}
\newacronym{ffn}{FFN}{feed-forward network}

\title{\LARGE \bf
A Point-Based Approach to Efficient LiDAR Multi-Task Perception
}

\author{Christopher Lang$^{1,2}$, Alexander Braun$^{1}$,
    Lars Schillingmann$^{1}$,  Abhinav Valada$^{2}$
    \thanks{$^{1}$Robert Bosch GmbH, Stuttgart, Germany}%
    \thanks{$^{2}$Department of Computer Science, University of Freiburg, Germany}%
}

\begin{document}

\maketitle
\thispagestyle{empty}
\pagestyle{empty}

\begin{abstract}
Multi-task perception networks hold great potential as they can improve performance and computational efficiency compared to their single-task counterparts, facilitating online deployment. 
However, current multi-task architectures in point cloud perception combine multiple task-specific point cloud representations, each requiring a separate feature encoder, making the network significantly large and slow.
In this work, we propose PAttFormer, an efficient multi-task learning architecture for joint semantic segmentation and object detection in point clouds, only relying on a point-based representation.
The network builds on transformer-based feature encoders using neighborhood attention and grid-pooling, complemented with a query-based detection decoder using a novel 3D deformable-attention detection head topology.
Unlike other LiDAR-based multi-task architectures, our proposed PAttFormer does not require separate feature encoders for multiple task-specific point cloud representations, resulting in a network that is 3$\times$ smaller and 1.4$\times$ faster while achieving competitive performance on the nuScenes and KITTI benchmarks for autonomous driving perception. 
We perform extensive evaluations that show substantial improvement from multi-task learning, achieving $+1.7\%$ in mIoU for LiDAR semantic segmentation and $+1.7\%$ in mAP for 3D object detection on the nuScenes benchmark compared to the single-task models.
\end{abstract}
    

\section{Introduction}
\label{sec:intro}

Predicting object-level bounding boxes and point-level semantic labels from LiDAR-based point clouds is a vital part of many autonomous driving systems, enabling them to navigate safely around objects and determine drivable areas.
Combining these two perception tasks in a single multi-task network architecture promises substantial benefits by sharing and jointly training parameters, including generalized latent representations that are less prone to overfitting and reduced latencies in the perception pipeline.
However, learning the dense segmentation and sparse detection tasks in a multi-task setting can also lead to competition for network parameters, resulting in unbalanced performances on single tasks. 
While the research on point cloud perception has consistently shown performance gains from multi-task learning~\cite{lidarmultinet,lidarformer,dwsnet}, the aforementioned challenges seem more dominant in the vision domain, where single-task networks often outperform their multi-task counterparts~\cite{vodisch2022continual,bevsic2022dynamic,vodisch2023codeps,mohan2022perceiving}.  
However, it remains unclear which characteristic of point cloud perception contributes to the performance improvement observed in multi-task training.

\begin{figure}
    \centering
    \footnotesize
    \includegraphics[width=0.95\linewidth]{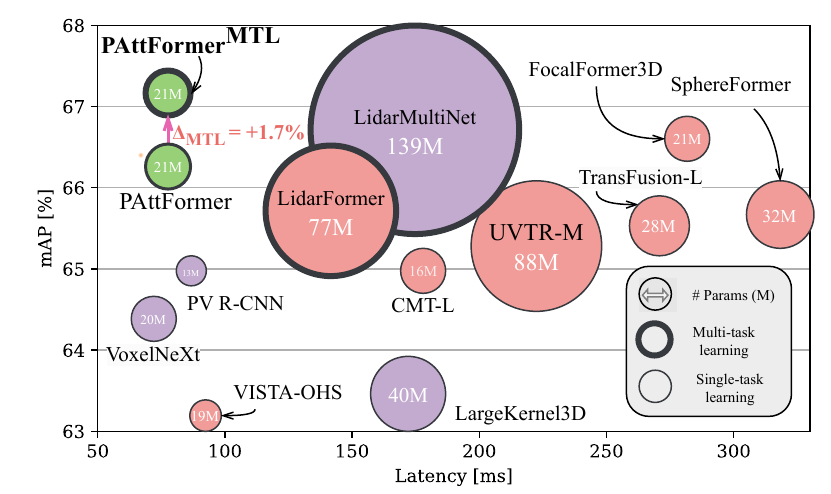}
    \caption{Performance-latency trade-off compared to related \gls{lidar} object detection models on the nuScenes validation set. Model sizes are encoded in the blob diameter.}
    \label{fig:latency-comparison}
    \vspace{-0.4cm}
\end{figure}

Many multi-task models in \acrshort{lidar} perception feature a separate point cloud representation for each task decoder~\cite{lidarmultinet,lidarformer,dwsnet}, e.g., the birds-eye view projection for 3D object detection or voxelization for semantic segmentation. 
This results in multi-task architectures that only share a small voxel encoder and feature heavy task-specific branches dedicated to the complementary point cloud representations, making these architectures parameter heavy and slow (refer to \figref{fig:latency-comparison} for a latency comparison). 
We are interested in whether architectures that share larger portions of parameters can attain similar performance improvement through multi-task learning or if they are prone to performance trade-offs in single tasks, as observed in the vision domain.



To investigate this hypothesis, we propose a simple and efficient point-based multi-task architecture that enables hard parameter-sharing in large parts of the model.
We build the network using transformer layers based on neighborhood attention to adapt to the local geometry of the 3D data and design a lightweight detection decoder head based on 3D deformable attention that can be trained end-to-end with a multi-task objective for 3D object detection and semantic segmentation.
We perform extensive evaluations on the \acrlong{nuscenes} and \acrlong{kitti} datasets, where our proposed multi-task architecture achieves competitive accuracy to state-of-the-art multi-task perception architectures while being  3$\times$ smaller and 1.4$\times$ faster.
During data efficiency experiments using limited annotated training data, we observed consistent performance improvement from multi-task learning for semantic segmentation and object detection compared to single-task training.

Our main contribution can be summarized as follows:
\begin{itemize}[topsep=0pt]
    \item A novel multi-task architecture that learns raw point cloud features without task-specific projections.
    \item A lightweight query-based detection head based on 3D deformable attention on point-based representations.
    \item Extensive evaluations of our method on the nuScenes and KITTI benchmarks for semantic segmentation and 3D object detection.
    \item Evaluations of data efficiency for both single-task and multi-task training.
\end{itemize}

\section{Related Work}
\label{sec:related_work}

In this section, we review \gls{lidar}-based semantic segmentation and object detection methods and embed our work in the landscape of multi-task architectures.

{\parskip=3pt
\noindent\textit{3D Semantic Segmentation}:
Early works adapted point-based methods~\cite{pointnet++,kpconv,hurtado2022semantic} from indoor domains to outdoor scenes.
However, the point aggregation imposes high computational efforts due to large-scale pointsets from \gls{lidar} scanners due to the neighbor search and decreasing point density with range. 
Voxel-based methods build on a structured representation of the point cloud that defines a neighborhood relationship, which allows adapting sparse 3D-convolutions~\cite{minkunet,cylinder3d}. They increase efficiency by trading off between the granularity of the voxel grid and memory requirements. 
However, the connectivity of feature representations in distant regions of the point cloud is challenging due to the small kernel size, which was approached by cylindrical voxel shapes~\cite{cylinder3d,polarnet} or improved kernel designs~\cite{link}.
In recent years, transformer-based modules have been proposed for enabling long-range connections in the feature extractor, either as plug-in modules to sparse convolution networks~\cite{sphericaltransformer} or as standalone architectures~\cite{ptv2,rangevit}.}


{\parskip=3pt
\noindent\textit{3D Object Detection}:
Point-based 3D detectors~\cite{3dssd,pointrcnn} take unstructured point clouds and process them using set-abstraction modules that aggregate point features on a set of candidate points during the downsampling process.
Although high accuracy is achieved in detecting single classes, voxel-based methods dominate multi-class detection benchmarks.
Such methods build on a pillar- or voxel-based feature extractor combined with a keypoint~\cite{centerpoint,sphericaltransformer} or anchor-based detection head~\cite{pvrcnn}.
Recently, transformer-based operators have drawn research attention due to their flexibility in capturing contextual features. For attention-based feature extractors, SWFormer~\cite{swformer} uses attention windows of constant volume and proposes to batch windows with a similar number of tokens to deal with varying point density, while other methods~\cite{flatformer,ptv2} explore the possibility of performing self-attention on windows of equal neighbor size but varying volumes.
In the detection head, attention-based methods employ features of candidates as detection query embeddings, which are refined in consecutive transformer decoder layers. 
Those candidates are extracted from center points in \gls{bev}-based methods~\cite{focalformer} or encoded 3D foreground points~\cite{li3detr}.}

 

{\parskip=3pt
\noindent\textit{Multi-Task LiDAR Perception}: 
\gls{lidar}-based multi-task methods in the literature limit hard-parameter sharing to a pillar-based~\cite{dwsnet} or voxel-based~\cite{lidarmultinet,lidarformer,lidarmtl} feature encoder.
Except for the pillar-based design in DWSNet~\cite{dwsnet}, which concentrates on detection performance, these multi-task architectures employ multiple decoder heads with task-specific point cloud representation and feature encoding. 
While~\cite{dwsnet,lidarmtl} limit the inter-dependency of tasks to the encoder, LidarMultiNet~\cite{lidarmultinet} proposes to re-project the encoded~\gls{bev} detection features back into the global voxel grid for a low-level feature exchange, which they observed to benefit the segmentation task. 
LidarFormer~\cite{lidarformer} further proposes a query-based exchange of task-specific prediction features in the task decoder, yielding a performance boost to the detection accuracy.
Contrary to~\cite{lidarmultinet} and~\cite{lidarformer}, we propose a multi-task architecture based on a point-based representation across all task heads. This eliminates the task-specific feature encoding and explicit feature-exchange mechanisms and allows for a more parameter and computationally efficient architecture at comparable benchmark performances.}

\section{Technical Approach}
\label{sec:system}

In this section, we describe our proposed multi-task architecture in \secref{sec:architecture}, followed by the design of the point attention operator in \secref{sec:patt}, the detection decoder head in \secref{sec:detection_head}, and finally our multi-task optimization approach in \secref{sec:mto}.

\begin{figure*}[ht]
    \centering
    \includegraphics[width=1\linewidth]{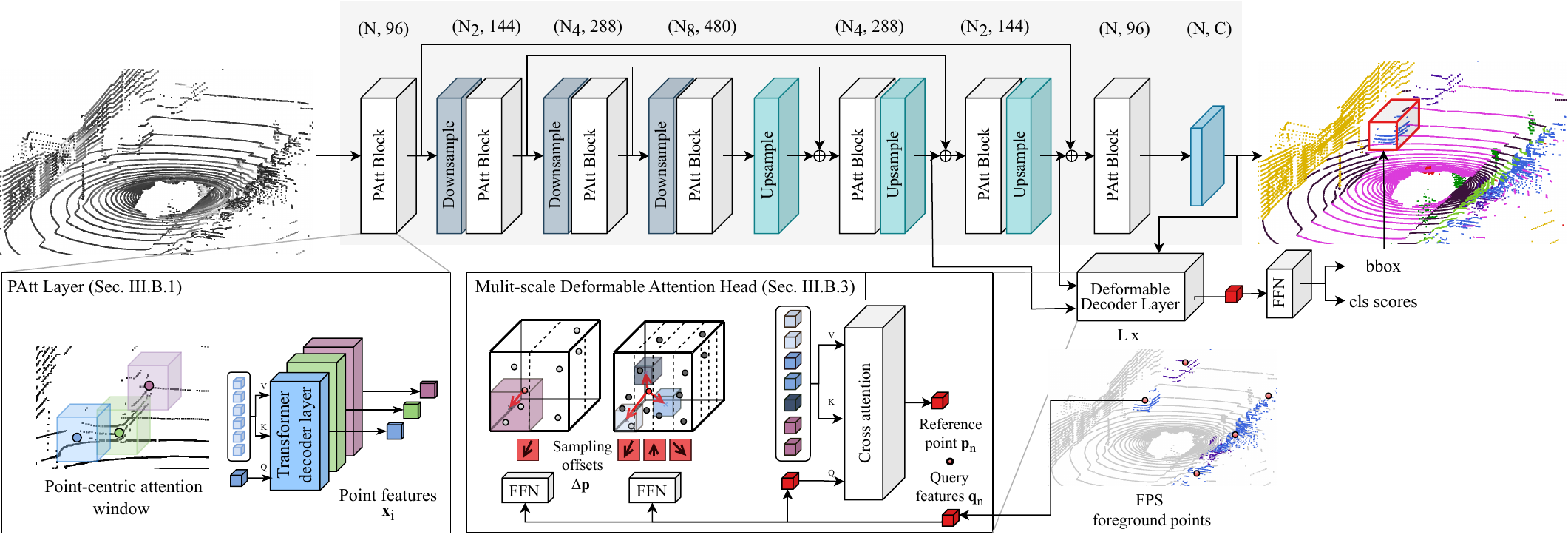}
    \caption{System architecture for a bottom-up set-up using the proposed PAtt modules that takes a raw point cloud as input and predicts a semantic class label for each scan point. In this setup, we start feature extraction on a subsampled point cloud with a larger attention window volume. We enrich the point cloud with previously masked points in the following stages while decreasing the attention window volume to balance computational complexity.}
    \label{fig:bottom_up_system_diagram}
    \vspace{-0.4cm}
\end{figure*}

\subsection{Problem Definition}

Our model takes as input a point cloud in the form of a bag of unordered points $\mathbf{P}$, representing a three-dimensional scene captured by range sensors such as a \gls{lidar} scanner. 
Each point $\mathbf{p}_i \in \mathbf{P}$ in the point cloud is characterized by its three-dimensional coordinates (x, y, z) and additional attributes such as intensity or color.
The model predicts a per-point semantic class label $s_i$, indicating the semantic class of the surface that reflected the laser scan, as well as a set of 3D-oriented bounding boxes $\mathbf{b}_m$ and a class label $c_m$ for all identified objects in a scene.
The method is trained with a dataset that contains semantic class and bounding box annotations for each frame.


\subsection{Multi-Task LiDAR PAttFormer Architecture}
\label{sec:architecture}
In a point cloud, the density and other attributes of points may not be uniform across different locations — in 3D scanning, such density variability might arise from perspective effects, radial density variations, motion, etc.
Therefore, many recent architectures use task-specific point cloud projection to regain the density of neighborhoods and use convolutional modules to extract features. 
To become task agnostic in our network design, we use a point-based representation, i.e., we only operate on the non-empty part of the point cloud described by its coordinates and point feature vector.
Our core novelty is a point-based multi-task perception model for simultaneous semantic segmentation and object detection. This model avoids task-specific projections and allows hard-parameter sharing in the feature encoding and decoding stages.
The model builds on transformer-based feature extraction and prediction modules that operate on point-wise representations, using the PAtt-layers described in \secref{sec:patt}.

We follow a multi-scale approach with a U-Net architecture depicted in \figref{fig:bottom_up_system_diagram}.
The consecutive encoder stages downsample the point density by $8\times$ to extract high-level point cloud features, which are consecutively upsampled and fused with skipped encoder features to preserve fine-grained properties. 
For downsampling, we use an efficient grid pooling method~\cite{ptv2} that maintains a distributed representation of the points by structuring the point cloud into a regular grid and max-pooling the point features while mean-pooling the point coordinates within each grid cell. 
The grid size increases with each stage, where $N_i$ denotes the number of non-empty grid cells at stage $i$. 
This avoids the expensive neighborhood search and guarantees an upper limit to the point density. 
During upsampling, point features are unmapped to all points in the grid cell and fused with the point features of the corresponding encoder stage.
Dense semantic segmentation labels are predicted by a lightweight task-head described in~\secref{sec:segmentation_head}.
Sparse detection predictions are generated by a query-based detection head using deformable attention on the foreground point features described in \secref{sec:detection_head}.




\subsubsection{Point-Based Attention (PAtt)}
\label{sec:patt}

Instead of the common PointNet architectures, we employ attention-based feature operators that embed positional information in a contextual relative encoding for point-based feature extraction.
The design is similar to~\cite{ptv2}, except that we replace the grouped vector attention with scaled-dot product attention for its lower latency.
We compute point-neighborhood windows $\mathcal{N}_i$ for each point separately using a ball-query approach, which selects up to $M$ neighboring points within a specified volume around a query point $\mathbf{p}_i$.
The ball query takes the form of a breadth-first search algorithm on voxel coordinates of each point to exploit the neighbor-aware representation in a voxel grid, similar to the voxel query proposed in~\cite{voxelset}.
Our ablation study shows that this approach can compete with the expensive nearest-neighbor search on the unordered point cloud used in~\cite{ptv2}.

We employ a contextual relative position encoding dependent on the query point feature $\mathbf{x}_{i}$ as a bias to the attention weights given by
\begin{align}
    b_{i,j} = \left( \mathbf{x}_i \mathbf{W}_{r} \right) \mathbf{r}_{i,j}^T  \quad \forall j \in \mathcal{N}_i,
\end{align}
where $\mathbf{r}_{i,j} = \phi_z(\mathbf{p}_i - \mathbf{p}_j) \in \mathcal{R}^{d_z}$ is the relative position encoding between query point $\mathbf{p}_i$ and neighbor point $\mathbf{p}_j$  by a two-layer MLP $\phi_z$ and $\mathbf{W}_r \in \mathcal{R}^{d_q \times d_z}$ is a learnable parameter matrix.
The neighborhood attention is then given by
\begin{align}
    \mathbf{y}_i = \sum_{j \in \mathcal{N}_i}  \sigma\left((\mathbf{W}_q\mathbf{x}_i) (\mathbf{W}_k\mathbf{x}_j)^T + b_{i,j} \right) \mathbf{W}i_v(\mathbf{x}_j),
\end{align}
where $\mathbf{W}_{\{q,k,v\}}$ are the query, key, and value projections and $\sigma$ denotes the softmax normalization function.

We integrate the point attention in a \gls{patt}-layer following a standard transformer layer with batch normalization. 
A \gls{patt} block consists of $L$ consecutive \gls{patt} layers with the same point-attention windows.

\subsubsection{Segmentation Head}
\label{sec:segmentation_head}
The segmentation head consists of an additional PAtt block operating on the full-resolution point cloud and a final classification layer that maps each point feature to the classification scores separately.

\subsubsection{Detection Head}
\label{sec:detection_head}

We design a query-based transformer decoder based on a 3D multi-scale formulation of deformable attention as depicted in \figref{fig:bottom_up_system_diagram}.
First, we extract $Q=200$ detection queries using the furthest point sampling method from the set of foreground points. We define foreground points with a predicted semantic class score of any detection object type exceeding a threshold $\delta_{fg} = 0.2$. In the detection-only setting, we train the semantic head with a pseudo foreground class label for all points inside a ground truth bounding box.

For the attention computation, the query point features $\mathbf{q}_k$ are mapped onto offset vectors $\Delta \mathbf{r}_k = \phi_{off}(\mathbf{q}_k)$ to learn context-dependent sampling points relative to the reference points $\mathbf{r}_k$.
Those sampling points are the center of a voxel query to accumulate point features that build the deformable attention window. 
We combine the point features of two scale levels, for which we each learn a separate set of offset vectors.
This approach is used in a multi-head deformable attention setting, where each head computes a separate set of sampling offsets. 
The query features are refined for $L$ decoder layers and finally fed into the classifier head \textit{FFN}, which regresses a bounding box and class scores, including a background class.

\subsection{Multi-Task Training}
\label{sec:mto}

We use point-wise cross-entropy classification loss $L_{cls,s}$ and a Lovasz Loss $L_{lov,s}$ for the segmentation task.
As detection losses, we use a binary focal loss  $L_{obj,d}$ for objectness prediction, box classification with a cross-entropy loss $L_{cls,d}$, center offset prediction with a smooth-L1 loss $L_{center,d}$, as well as size prediction with a smooth-L1 loss $L_{size,d}$.
In the single-task setting, we compensate for the semantic segmentation losses with a point-wise focal loss for foreground classification. We use uncertainty weighting~\cite{uncertaintyweighting} to balance optimizing the multi-task loss terms. 



\section{Experimental Evaluation}
\label{sec:experiments}

\begin{table}[t]
\footnotesize
\centering
\caption{Training settings. All angles are in degrees.}
\label{tab:train_hyperparameters}
\begin{tabular}{lll} 
\toprule
Config           & nuScenes  value    & KITTI  value                                                              \\ \midrule
Optimizer        & AdamW                                                                               & AdamW                                                                               \\
Learning rate    & $1e-4$                                                                              & $1e-4$                                                                              \\
LR schedule      & CosineAnnealingLR                                                                   & CosineAnnealingLR                                                                   \\
Weight decay     & $1e-2$                                                                              & $1e-2$                                                                              \\
Epochs           & 36                                                                                  & 50                                                                                  \\
Pointcloud range & \begin{tabular}[t]{@{}l@{}}min=[-50, -50, -5] \\ max=[50, 50, 3]\end{tabular} & \begin{tabular}[t]{@{}l@{}}min=[0, -40, -3] \\ max=[70.4, 40, 1]\end{tabular} \\
Random scale     & [0.95, 1.05]                                                                        & [0.95, 1.05]                                                                        \\
Random rotate    & [-45, 45]                                                                           & [-45, 45]                                                                           \\
Random flip      & p=0.5                                                                             & p=0.5                                                                             \\
Laser Mix        & pitch angles=[-25, 3]                                                               & pitch angles=[-25, 3]                                                               \\
Polar Mix        & \begin{tabular}[t]{@{}l@{}}swap ratio=0.5\\rotate paste ratio=1.0 \end{tabular} & \begin{tabular}[t]{@{}l@{}}swap ratio=0.5\\rotate paste ratio=1.0 
\end{tabular}  \\ \bottomrule                                           
\end{tabular}
\vspace{-0.5cm}
\end{table}

In this section, we evaluate our approach on the nuScenes and KITTIdatasets and compare the results against related state-of-the-art methods in~\secref{sec:experiments_detection} for object detection and in~\secref{sec:experiments_segmentation} for semantic segmentation.
We then analyze the effects of multi-task learning in \secref{sec:multi-task} and various architecture design choices~\secref{sec:experiments_ablation} and report the results of our data efficiency experiments in~\secref{sec:experiments_data_efficiency}. Finally, we apply the PAttFormer to odometry estimation in \secref{sec:odometry} and present qualitative predictions in~\secref{sec:demos}.

\subsection{Datasets}
\label{sec:datasets}
We evaluate our proposed method on driving domains on the large-scale KITTI and nuScenes datasets that provide the recordings of a single, roof-mounted, \gls{lidar} scanner's positional and radiometric measurements.

\paragraph{nuScenes}
The \acrlong{nuscenes} dataset consists of 1000 driving sequences of approximately 20 seconds with a \gls{lidar} frequency of $20Hz$. The sequences are split into sets of 700, 150, and 150 for training, validation, and testing and are annotated for 10 classes with a long-tail distribution. 
The samples were recorded with a Velodyne HDL-32E rotating laser scanner, capturing approximately 30k points per frame~\cite{panopticnuscenes}.
For object detection, annotations are available at $2 Hz$, and the point clouds are commonly densified by accumulating over the previous and subsequent sweeps. 
This dataset suffers from class imbalance, which makes the detection task more challenging.

\paragraph{SemanticKITTI}
The \acrlong{semantickitti} dataset contains 23k training samples and 20k testing samples in autonomous driving scenes with panoptic segmentation annotations for 8 \textit{thing} and 12 \textit{stuff} classes.
The samples were recorded with a Velodyne HDL-64E rotating laser scanner, capturing approximately 100k points per cycle~\cite{semantickitti}. 
We generate enclosing bounding boxes for all \textit{thing} class types for multi-task learning and follow the common practice to validate the performance on \textit{sequence 08}.
We additionally provide detection performance evaluation on the \acrlong{kitti} dataset for benchmark metrics of AP$_{3D}$ at 40 recall points for the car class using an IoU threshold of 0.7 for matching with ground truth objects.

\subsection{Implementation Details}

Our implementation is primarily built on the MMDetection3D~\cite{mmdet3d} codebase. 
The training parameters are detailed in \tabref{tab:train_hyperparameters}. 
We perform our experiments on a dual-GPU setup using RTX 3090. The reported frame rates are measured on a single A100 for comparability with ~\cite{lidarmultinet,lidarformer} using the MMDetection3D implementations wherever available.
We report the performance using mean results across all classes on the nuScenes and SemanticKITTI datasets. The PAttformer model parameters are initialized with self-supervised pre-training on the respective train set using the TARL method~\cite{tarl}.
We also report results for an efficient multi-task baseline method, which we call MinkUCP, using a MinkowskiUNet~\cite{minkunet} segmentation network with a CenterPoint detection head~\cite{centerpoint} with the same loss terms as our proposed PAttFormer model.
We use single sweeps for evaluation and train on accumulated point clouds of three scans for multi-task learning by ignoring unlabeled points in the segmentation losses. For detection training, we follow~\cite{focalformer,querycontrast} to initialize additional detection queries from noisy ground truth boxes for faster convergence. 
We further use class-balanced sampling~\cite{classbalanced_sampling} augmentation during the first 10 training epochs.

\subsection{3D Object Detection Results}
\label{sec:experiments_detection}

\begin{table*}[t]
\footnotesize
\centering
\caption{Detection results on the nuScenes and KITTI benchmarks. Models marked with (\multitask) were trained in a multi-task setting. Kitti AP is reported for class "Car" at 40 recall points. Frame rates marked with (*) were taken from the cited references.}
\label{tab:det-benchmark}
\begin{tabular}{@{}llllllll|lllll@{}}
\toprule
Methods                                  & Representation & \multicolumn{2}{c}{nuScenes [val]} & \multicolumn{4}{l}{nuScenes [test]} & \multicolumn{5}{l}{KITTI [val] $\text{AP}_{3D}$}   \\
                                         &                & mAP              & NDS             & mAP    & NDS   & Params & FPS       & Easy & Mod. & Hard & Params & FPS  \\ \midrule
PointRCNN \cite{pointrcnn}               & P              & -                & -               & -      & -     & -      & -         & 88.9 & 78.6 & 77.4 & 4.0M     & 8.7    \\
3D-SSD \cite{3dssd}                      & P              & 56.4             & 42.7            & -      & -     & 2.5    & 18.9      & 89.7 & 79.4 & \underline{78.7} & 2.5M  & 18.6 \\
PV-RCNN \cite{pvrcnn}                    & V+P            & 65.0             & 53.6            & -      & -     & 13.2 & 10.5   & \underline{90.3} & \underline{81.4} & 76.8 & 13.1M  & 13.1 \\
CenterPoint  \cite{centerpoint}          & V+BEV          & 57.1             & 65.4            & 58.0   & 65.5  & 8.9    & 17.8      & -    & -    & -    & -     & -    \\
Li3DDeTr \cite{li3detr} & V+BEV          & -                & -               & 61.3   & 67.6  & - & -    & 87.6 & 76.8 & 73.9 & -     & -    \\
SphereFormer \cite{sphericaltransformer} & V+BEV          & -                & -               & 65.5   & 70.7  & - & -    & -    & -    & -    & -     & -    \\
LiDARMultiNet\multitask~\cite{lidarmultinet}& V+BEV        & 63.8            & 69.5            & 67.0   & 71.6  &139.0M & 7.9*   & -    & -    & -    & -     & -    \\
LiDARFormer\multitask~\cite{lidarformer}  & V+BEV         & \underline{66.6} & \underline{70.8}            & \textbf{68.9}   & \underline{72.4}  & 77.0M & 6.1*    & -    & -    & -    & -     & -    \\ 
FocalFormer3D \cite{focalformer}         & V+BEV          & 66.5 & \textbf{71.1}            & \underline{68.7}   & \textbf{72.6}  & 21.3M  & 3.6       & -    & -    & -    & -     & -    \\ \midrule
PAttFormer                               & P              & 64.3            & 68.7           & 65.3    & 69.2  & 21.5M  & 11.0      & 88.8 & 78.5 & 77.3 & 21.5M & 8.4    \\
PAttFormer\multitask                     & P              & \textbf{66.7}   & 69.5            & 67.0    & 69.3  & 21.5M  & 11.0      & \textbf{91.0} & \textbf{82.0} & \textbf{79.1} & 21.5M     & 8.4    \\ \bottomrule
\end{tabular}
\end{table*}
\begin{table*}[ht]
\footnotesize
\centering
\caption{Semantic segmentation results on nuScenes and SemanticKITTI benchmarks. Models marked with (\multitask) were trained in a multi-task setting. Frame rates marked with (*) were taken from the cited references.}
\label{tab:semseg-benchmark}
\begin{tabular}{@{}llllll|llll@{}}
\toprule
                                         &                & \multicolumn{4}{l}{nuScenes}     & \multicolumn{4}{l}{SemanticKITTI}  \\
Method                                   & Representation & mIoU [val] & mIoU [test] & Params & FPS      & mIoU[val] & mIoU [test] & Params & FPS      \\  \midrule
\multicolumn{9}{l}{\textbf{Convolutional methods}}                                                                                         \\
PolarNet \cite{polarnet}                 & V              & 71.0       & 69.8        & 14.0M &  14.9 & 53.6   & 54.3 & 14.0M    & 12.1        \\
MinkowskiNet \cite{minkunet}& V              & 70.3       & -           & 37.8 &  21.7 & 61.1   & 63.1 & 21.7     & 15.3        \\
Cylinder3D \cite{cylinder3d}             & V              & 76.1       & 77.2        & 55.8M &  8.9 & 64.3   & 61.8 & 55M.8  & 8.9      \\
LiDARMultiNet\multitask~\cite{lidarmultinet}& V           & \underline{82.0}       & \textbf{81.4}        & 139.0M     & 7.9*    & -     & -    & -     & -        \\
LiDARFormer\multitask~\cite{lidarformer}    & V           & \textbf{82.7}       & \underline{81.0}        & 77.0M     & 6.1*        & -     & -    & -        \\
\multicolumn{9}{l}{\textbf{Transformer-based methods}}                                                                                     \\
RangeViT \cite{rangeformer}              & RV             & 75.2      & -           & 27.1M & -        & -     & 64.0  & 27.1M     & -        \\
SphereFormer \cite{sphericaltransformer} & V              & 78.4      & 81.9        & 32.3M & 5.4     & 67.8        & \textbf{74.8} & 32.2M & 2.6      \\ \midrule
PAttFormer                         & P              & 78.0      & 76.9           & 16.9M & 14.8     & \underline{68.4} & 67.3     & 16.9M & 9.5      \\
PAttFormer\multitask                & P              & 79.8      & 78.7          & 16.9M & 14.8     & \textbf{69.8} & \underline{69.3}     & 16.9M & 9.5      \\ \bottomrule
\end{tabular}
\end{table*}
\begin{table*}[ht]
\footnotesize
\centering
\caption{Per-class semantic segmentation results on nuScenes val set.}
\label{tab:nuscenes_val_semseg_per_class}
\begin{tabular}{l|l|lllllllllllllll} \toprule
Method       & mIoU & 
\rot{barrier} & 
\rot{bicycle} & 
\rot{bus}  & 
\rot{car}  & 
\rot{construction} & 
\rot{motorcycle} & 
\rot{pedestrian} & 
\rot{traffic cone} & 
\rot{trailer} & 
\rot{truck} & 
\rot{driveable} & 
\rot{other flat} & 
\rot{sidewalk} & 
\rot{terrain} & 
\rot{manmade}  \\ \midrule
PolarNet~\cite{polarnet}     & 71.0 & 74.7    & 28.2    & 85.3 & 90.9 & 35.1         & 77.5       & 71.3       & 58.8         & 57.4    & 76.1  & 96.5      & 71.1       & 74.7     & 74.0    & 87.3    \\
Cylinder3D~\cite{cylinder3d}   & 76.1 & 76.4    & 40.3    & 91.2 & \underline{93.8} & 31.3         & 78.0       & 78.9       & 64.9         & 62.1    & 84.4  & 96.8      & 71.6       & \textbf{76.4}     & 75.4    & 90.5    \\
PVKD~\cite{pvkd}         & 76.0 & 76.2    & 40.0    & 90.2 & \textbf{94.0} & 50.9         & 77.4       & 78.8       & 64.7         & 62.0    & 84.1  & 96.6      & 71.4       &\textbf{76.4}     & 76.3    & 90.3    \\
RPVNet~\cite{rpvnet}       & 77.6 & \textbf{78.2}    & 43.4    & 92.7 & 93.2 & 49.0         & 85.7       & 80.5       & \underline{66.0}         & 66.9    & 84.0  & \underline{96.9}      & 73.5       & 75.9     & 76.0    & \underline{90.6}    \\
SphereFormer~\cite{sphericaltransformer} & \underline{78.4} & 77.7    & 43.8    & 94.5 & 93.1 & 52.4         & \underline{86.9}       & \underline{81.2}       & 65.4         & 85.3  & \textbf{97.0}      & \underline{73.4}       & 75.4     & 75.0    & \textbf{91.0}   \\  \midrule    
PAttFormer 	 & 78.0 & 76.1 	  & \underline{48.5} 	& \underline{95.2} & 91.2 & \underline{58.2} 		 & 86.7 	  & 80.0 	   & 63.0 		  & 72.3 	& \underline{87.1}  & 96.3 		& 73.0 		 & 74.3 	& 74.0 	 & 88.1 	\\ 
PAttFormer \multitask	 
			& \textbf{79.8}  & \underline{77.8}    & \textbf{50.8}    & \textbf{95.5} & 92.6 & \textbf{60.3}         & \textbf{88.3}       & \textbf{82.5}       & \textbf{67.0}         & \textbf{75.5}    & \textbf{88.8}  & 96.7      & \textbf{74.9 }      & 75.7     & 75.1   & 88.8   
			  \\ \bottomrule       
\end{tabular}
\vspace{-0.3cm}
\end{table*}

\tabref{tab:det-benchmark} summarizes the performance of our proposed architecture on the semi-dense perception task of 3D object detection on the \acrlong{kitti} val and \acrlong{nuscenes} test benchmarks.
During the post-processing stage, we select bounding boxes with confidence scores larger than 0.2, and we utilize \gls{nms} with an IoU threshold of 0.4.
We selected related multi-task architectures (marked with~\multitask) and alternative point-based, voxel-based, and center-based (\gls{bev}) methods with comparable model sizes or frame rates as baselines.
Our proposed PAttFormer model achieves $67.0\%$ in mAP on the nuScenes test set, outperforming all other methods with frame rates $>10Hz$. 
PattFormer even achieves comparable detection accuracy with related multi-task networks~\cite{lidarmultinet,lidarformer} at less than one-third of the parameter count. 
On the KITTI val benchmark, our model outperforms other point-based and two-stage methods by $+1.6\%$ and $+0.6\%$ in mAP on the \textit{car} class for moderate difficulty and achieves a frame rate of $8.4$Hz for a point cloud with 64 horizontal scanlines.

\subsection{Semantic Segmentation Results}
\label{sec:experiments_segmentation}

We evaluate the semantic segmentation performance on the \acrshort{nuscenes} and \acrshort{semkitti} in \tabref{tab:semseg-benchmark} benchmarks. 
We report the latency with the detection head excluded.
The proposed PAttFormer\multitask model achieves $78.7\%$ in mIoU on the nuScenes test set, outperforming other voxel-based transformer methods and in close range to the much larger multi-task networks. 
The per-class evaluation in \tabref{tab:nuscenes_val_semseg_per_class} suggests that the PAttFormer model exhibits especially strong performance on \textit{thing} classes, with improvement of $+2.5\%$ in IoU for pedestrians from multi-task training alone.
on the SemanticKITTI test set, the PAttFormer\multitask method achieves $69.3\%$ in mIoU without using test-time augmentations.
We observe a performance improvement through multi-task training of $+1.8\%$ and $+2.0\%$ in mIoU on the nuScenes and SemanticKITTI test sets, respectively.

\subsection{Multi-Task Learning Evaluation}
\label{sec:multi-task}

\begin{table*}[t]
\footnotesize
\centering
\caption{Ablation experiments on nuScenes validation set.
Detection is also split into \b{c}lose range ($\leq10m$), \b{m}edium range ($10m < x \leq30m$), and  \b{f}ar range ($30m < x$) to the GT bbox center.
}
\label{tab:ablation}
\begin{tabular}{@{}lllllllllll@{}}
\toprule
\multicolumn{2}{l}{Windowing} & Sampling & Loss balancing & Det. head   & Seq           & \multicolumn{4}{c}{Det}                                       & FPS  \\ 
Search         & Size         &          &              &             & mIoU          & mAP           & $AP_{c}$      & $AP_{m}$      & $AP_{f}$      &      \\ \midrule
VQ             & 32           & Grid     & Uncertainty  & DDETR3D     & \underline{79.8} & \textbf{66.8} & \underline{79.1} & \textbf{71.1} & \textbf{37.0} & 11.0 \\ \midrule
VQ             & 16           & Grid     & Uncertainty  & DDETR3D     & 74.3          & 52.6          & 73.4          & 58.0          & 24.9          & 10.7  \\
VQ             & 64           & Grid     & Uncertainty  & DDETR3D     & 79.5          & 64.5          & 76.6          & 66.4          & 32.1          & 9.2  \\
kNN            & 32           & Grid     & Uncertainty  & DDETR3D     & \textbf{79.9} & \underline{66.3} & \underline{79.1} & \underline{70.7} & \underline{36.5} & 8.3  \\
VQ             & 32           & None     & Uncertainty  & DDETR3D     & 75.8          & 48.7          & 70.7          & 53.7          & 22.0          & 7.0  \\
VQ             & 32           & Grid     & Uncertainty           & CenterPoint & 76.5          & 59.6          & 75.9          & 64.4          & 30.9          & 12.5 \\ 
VQ             & 32           & Grid     & GradNorm     & DDETR3D     & 79.3          & 64.8          & \textbf{79.3} & 67.2          & 30.6          & 11.0 \\ \bottomrule
\end{tabular}
\vspace{-0.3cm}
\end{table*}
\begin{table}[t]
\footnotesize
\centering
\caption{Multi-task evaluation on nuScenes validation set. Frame rates marked with $^\ast$ were taken from the cited references.}
\label{tab:multitask_nus}
\begin{tabular}{p{1.7cm}p{0.3cm}|p{0.4cm}p{0.6cm}|p{0.4cm}p{0.6cm}|p{0.3cm}p{0.7cm}}
\toprule
                                    & \multitask          & Segmentation  &          & Detection  &      & FPS          & Params \\ 
                                    &              & mIoU & $\Delta_{\text{MTL}}$   & mAP  & $\Delta_{\text{MTL}}$ &              &                \\ \midrule
LidarMultiNet~\cite{lidarmultinet}  & \ding{51}   & 82.0 & -        & \textbf{67.0} & -     & 7.9$^\ast$         & 139.0M    \\
LidarFormer~\cite{lidarformer}      & \ding{55}   & \underline{82.1} & -        & 65.9 & -     &         & 77.0M        \\
                                    & \ding{51}   & \textbf{82.6} & +0.5     & 66.0 & +0.1  & 6.1$^\ast$         & 77.0M        \\ 
MinkUCP                             & \ding{55}   & 74.5 & -        & 57.1 & -     & 13.7    & 44.1M        \\
                                    & \ding{51}   & 75.0 & +0.5     & 56.9 & -0.2  & 13.7    & 44.1M        \\ \midrule
PAttFormer                          & \ding{55}   & 78.0 &  -       & 64.3 & -     & 11.0    & 21.5M        \\
                                    & \ding{51}   & 79.8 & \textbf{+1.8}     & \underline{66.7} & \textbf{+2.4}  & 11.0      & 21.5M        \\ \bottomrule
\end{tabular}
\vspace{-0.4cm}
\end{table}

In this section, we evaluate the benefits of multi-task training for benchmark performance on the nuScenes dataset in \tabref{tab:multitask_nus}. Apart from models from the literature, we report results for a runtime-efficient baseline consisting of a MinkUNet~\cite{minkunet} voxel-encoder and a CenterPoint~\cite{centerpoint} detection head, which we call MinkUCP. This baseline model was trained with comparable training settings to our PAttFormer model, adapting the learning to $lr=0.01$.
PAttFormer runs at 1.4$\times$ faster frame rate compared to LiDARMultiNet~\cite{lidarmultinet} at slightly inferior performance by $-0.3\%$/$-2.2\%$ for the detection and semantic segmentation, respectively.
Our proposed point-based PAttFormer model outperforms the efficient MinkUCP baseline in segmentation and detection but with a slightly lower frame rate.
Note that the PAttFormer model achieves the highest improvement from multi-task learning among all architecture designs, with a $ +2.4\%$ increase in mAP for the detection task.

\subsection{Ablation Study}
\label{sec:experiments_ablation}

The proposed feature extractor depends on a multitude of hyperparameters. 
In \tabref{tab:ablation}, we summarize the effect of each hyperparameter on perception performance for models trained in a multi-task learning setting.

{\parskip=3pt \noindent\textit{Attention Window Size}}
First, we are interested in how the size of the attention window affects the model's ability to understand the scene.
We trained three models with attention windows of 16, 32, and 64 neighbor points, respectively, and achieved the best performance for a window size of 32. Interestingly, the larger window size 64 has lower detection performance, especially for distant objects, with the frame rate dropping by $-2Hz$ compared to computing attention over 32 neighbor points. We assume that the larger window size, especially in downsampled representations, includes points of multiple objects or background points that hinder learning discriminative features.
Next, we replaced the ball-query neighbor-search with a k-nearest neighbor algorithm, which had a minor effect on the model performance but caused the frame rate to drop by $-2.7Hz$.

{\parskip=3pt \noindent\textit{Multi-Scale Hierarchy}}
When disabling the point cloud downsampling in the encoder blocks, we observe a performance decrease by $-11.1\%$ in detection mAP. We assume the downsampling facilitates the connectivity of the point cloud, as it increases the field of view of each point with a fixed neighborhood size, allowing the model to learn more discriminative features in the encoder blocks.

{\parskip=3pt \noindent\textit{Loss Weighting}}
We compared two loss balancing strategies to trade-off performance across all tasks, \ie uncertainty weighting~\cite{uncertaintyweighting} and gradient normalization.
The uncertainty weighting strategy outperformed gradient normalization in both detection and segmentation tasks, whereby the performance drops most noticeably for the detection task by $-2.0\%$ in mAP.

{\parskip=3pt \noindent\textit{Detector Head}}
Lastly, we replace the detection head with a \gls{bev} projection and a CenterPoint decoder. Our results indicate that switching to a \gls{bev}-based detection head impairs detection ($-7.2\%$ in mAP) and segmentation ($-3.3\%$ in mIoU) performance alike. 
We suspect that the \gls{bev} projection and separate feature encoding required by the CenterPoint~\cite{centerpoint} method results in the decoupling of the perception tasks that impede multi-task learning. 
As a result, the optimizer struggles with tradingoff performance between both tasks such that the segmentation score even falls below the single-task performance by $-1.5\%$ in mIoU (compared to \tabref{tab:semseg-benchmark}). 
\subsection{Data Efficieny Analysis}
\label{sec:experiments_data_efficiency}
\begin{table}[t]
\footnotesize
\centering
\caption{Data efficiency results on nuScenes validation set.}
\label{tab:data-efficiency}
\begin{tabular}{lp{0.4cm}p{0.4cm}p{0.4cm}p{0.4cm}p{0.4cm}p{0.5cm}}
\toprule
 & \multicolumn{3}{c}{Seg mIoU [\%]} & \multicolumn{3}{c}{Det mAP [\%]} \\ 

\diagbox[dir=NW]{Model}{Train seq.} & 25\%      & 50\%      & 100\%     & 25\%      & 50\%      & 100\%     \\ \midrule
MinkowskiNet       & 61.9       & 70.0       & 70.3       & -          & -          & -          \\
CenterPoint        & -          & -          & -          &  41.2      & 52.8       & 57.1       \\
MinkUCP \multitask & 62.5       & 70.5      & 70.8        &  42.7      & 52.5       & 56.9       \\
PAttFormer         & \underline{65.0}       & \underline{74.8}       & \underline{78.0}       &  \underline{48.1}      & \underline{59.1}       & \underline{64.3}       \\
PAttFormer \multitask  & \textbf{70.9}   & \textbf{76.3}       & \textbf{79.8}       &  \textbf{52.6}      & \textbf{63.2}       & \textbf{66.7}       \\ \bottomrule
\end{tabular}
\vspace{-0.4cm}
\end{table}
\begin{figure*}[t]
    \centering
    \footnotesize
    \begin{minipage}[c]{0.87\textwidth}
        \centering
        \begin{minipage}[b]{\textwidth}
                \centering
            \begin{minipage}{0.48\textwidth}
                \includegraphics[width=\textwidth]{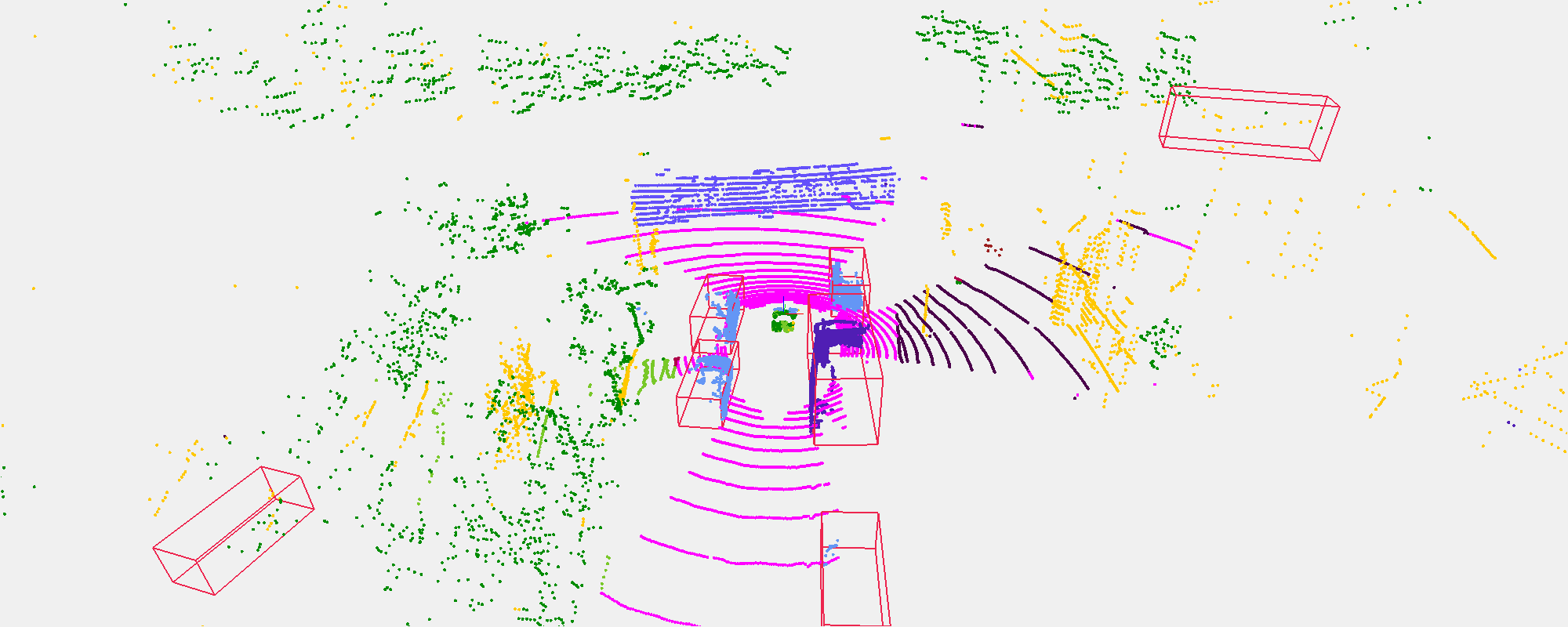}
            \end{minipage}%
            \hspace{0.01\textwidth}%
            \begin{minipage}{0.48\textwidth}\centering
                \includegraphics[width=\textwidth]{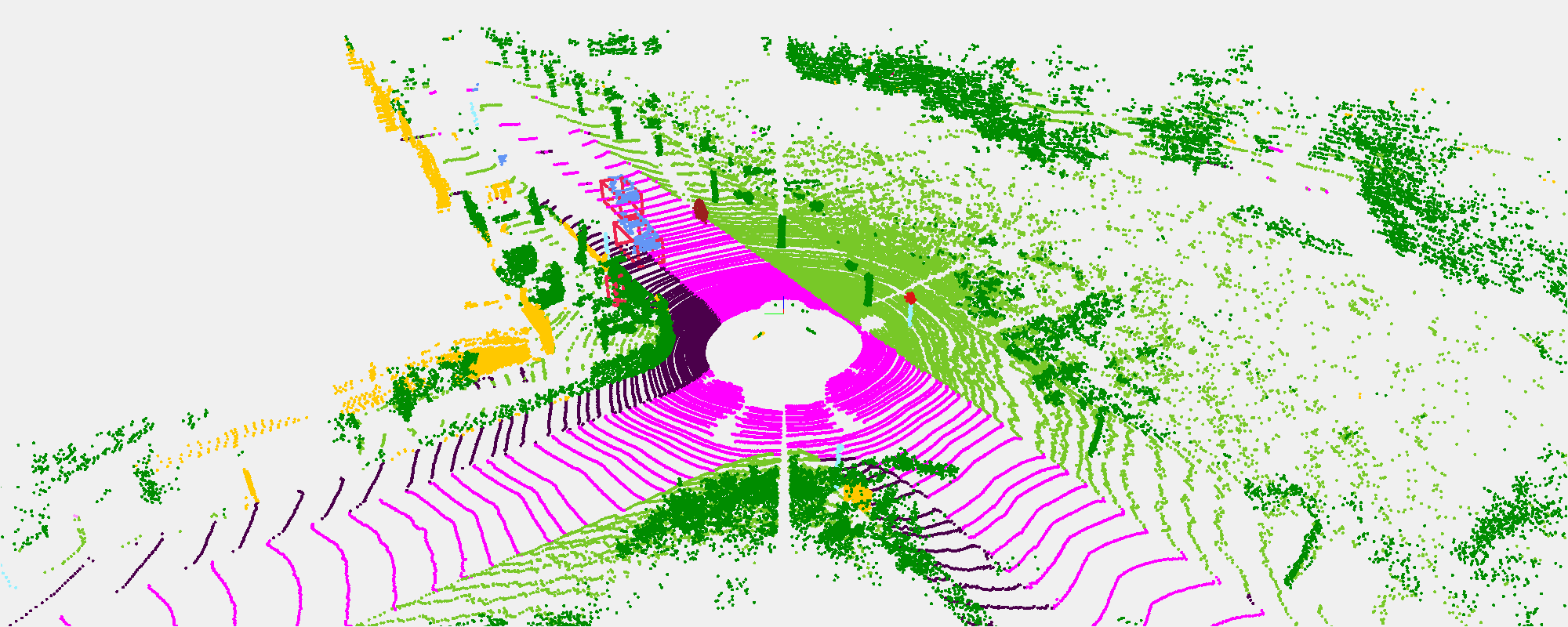}
            \end{minipage}\\[1mm]{a) \textit{PAtt} single-task network inferences.}
        \end{minipage}
        \vspace{1mm}
        
        \begin{minipage}[b]{\textwidth}
            \centering
            \begin{minipage}{0.48\textwidth}
                \includegraphics[width=\textwidth]{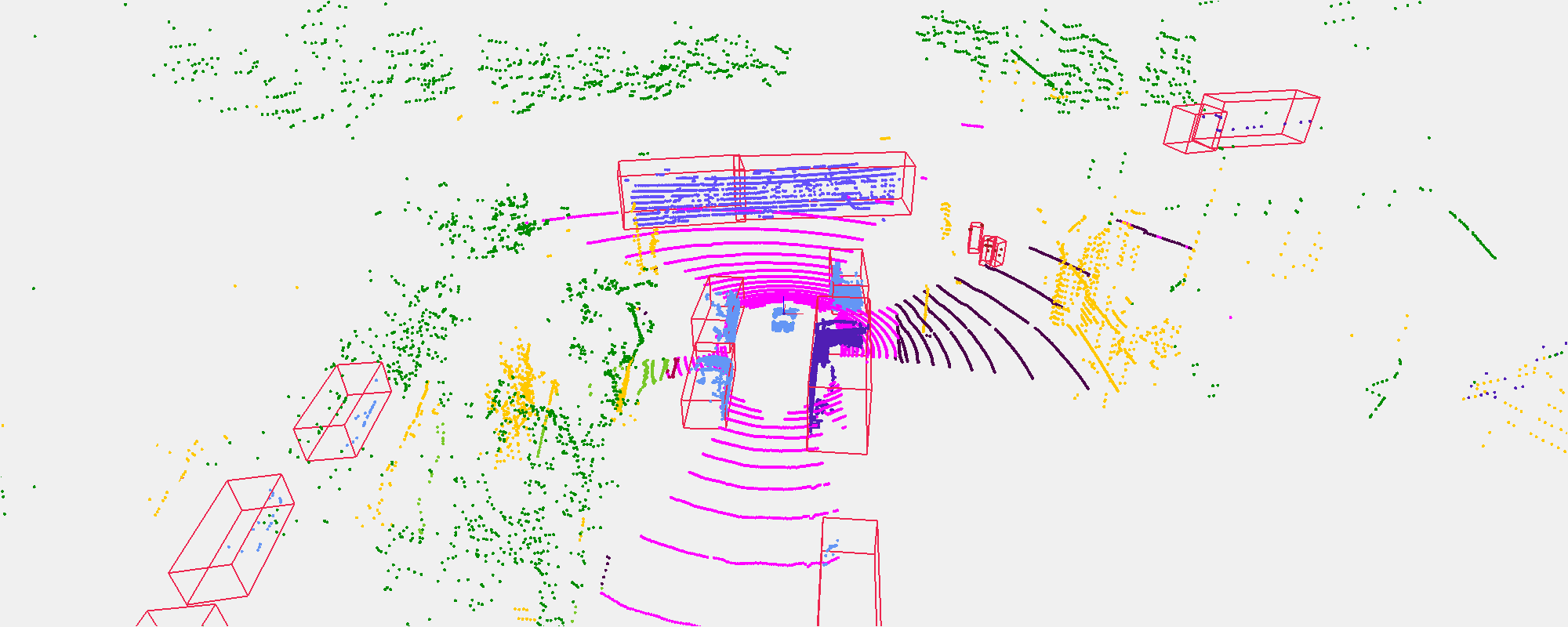}
            \end{minipage}%
            \hspace{0.01\textwidth}%
            \begin{minipage}{0.48\textwidth}\centering
                \includegraphics[width=\textwidth]{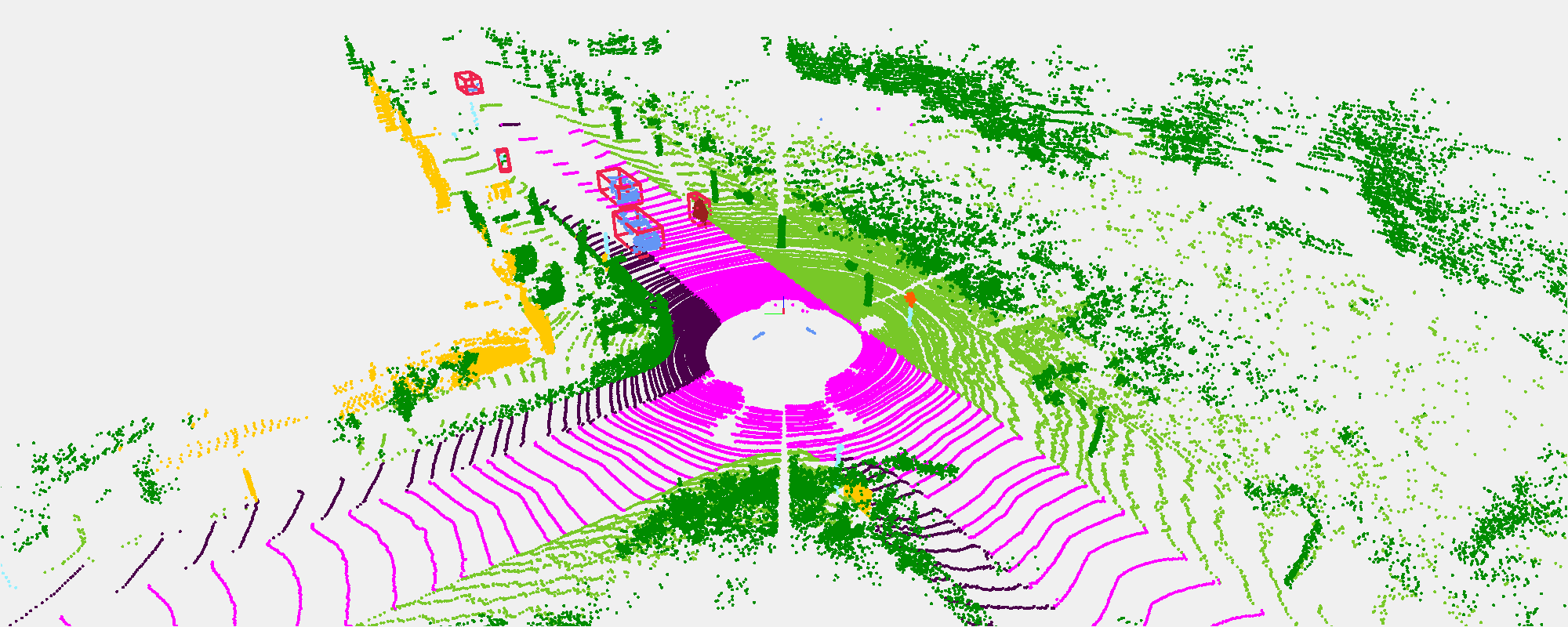}
            \end{minipage}\\[1mm]{b) \textit{PAtt} multi-task network inference.}
        \end{minipage}
    \end{minipage}
    \hspace*{\fill}
    \begin{minipage}[c]{0.12\textwidth}
        \scriptsize
        \begin{itemize}
            \setlength\itemsep{2pt}
                \item[\textcolor{barrier}{\Large\textbullet}] barrier
                \item[\textcolor{bicycle}{\Large\textbullet}] bicycle
                \item[\textcolor{bus}{\Large\textbullet}] bus
                \item[\textcolor{car}{\Large\textbullet}] car
                \item[\textcolor{construction-vehicle}{\Large\textbullet}] constr. veh.
                \item[\textcolor{motorcycle}{\Large\textbullet}] motorcycle
                \item[\textcolor{pedestrian}{\Large\textbullet}] pedestrian
                \item[\textcolor{traffic-cone}{\Large\textbullet}] traffic sign
                \item[\textcolor{trailer}{\Large\textbullet}] trailer
                \item[\textcolor{truck}{\Large\textbullet}] truck
                \item[\textcolor{driveable-surface}{\Large\textbullet}] driveable
                \item[\textcolor{other-flat}{\Large\textbullet}] other flat
                \item[\textcolor{sidewalk}{\Large\textbullet}] sidewalk
                \item[\textcolor{terrain}{\Large\textbullet}] terrain
                \item[\textcolor{manmade}{\Large\textbullet}] manmade
                \item[\textcolor{vegetation}{\Large\textbullet}] vegetation
                \item[\textcolor{black}{\Large\textbullet}] noise
        \end{itemize}
    \end{minipage}

    \caption{Comparison of predicted bounding boxes and segmentation labels on the 32-line scan from \gls{nuscenes} (left)  and 64-line scan from SemanticKITTI (right) val sets.}
    \label{fig:qualitative_example}
    \vspace{-0.3cm}
\end{figure*}

We assess the effect of multi-task training on overfitting by a data efficiency analysis.
Therefore, we train the models with multiple fractions of nuScenes \textit{train} set sequences and compare their performance on the \textit{val} set in \tabref{tab:data-efficiency}.
We compare validation set performance for training on 25\%, 50\%, and 100\% of training sequences, whereby we adopt the number of training epochs to 144, 72, and 36, respectively.
We also provide results for the MinkUCP multi-task baseline and its task-specific components.
The PAttFormer\multitask~model trained in a multi-task setting achieves good performance even with reducing training data variety to approximately 25\%. Using only half of the training annotations, the PAttFormer\multitask achieves $76.3\%$ in mIoU and $63.2\%$ in mAP, which is only $-1.7\%$/$-1.1\%$ less than the PAttFormer model trained on single-tasks with all available training annotations.
We did not observe such a trend with the MinkUCP baseline.



\subsection{Odometry Analysis}
\label{sec:odometry}

\begin{table}[t]
\footnotesize
\centering
\caption{Absolute Trajectory Error (ATE) and Absolute Rotational Error (ARE) averaged over all sequences for LiDAR odometry. }
\label{tab:odometry}
\begin{tabular}{@{}lcccc@{}}
\toprule
                  & \multicolumn{2}{c}{SemanticKITTI {[}Seq 8{]}} & \multicolumn{2}{c}{nuScenes {[}val{]}} \\
                  & ATE {[}m{]}     & ARE {[}rad{]}     & ATE {[}m{]}   & ARE {[}rad{]}  \\  \midrule
Raw pointcloud    & 5.97            & \underline{0.026}           & 1.80          & 0.018        \\
MinkowskiUNet            & 6.47            & 0.028           & \underline{1.56}          & \underline{0.013}        \\
PAttFormer           & \textbf{5.87}            & \underline{0.026}           & 1.61          & 0.014        \\
PAttFormer \multitask & \underline{5.88}            & \underline{0.026}           & \textbf{1.42}          & \textbf{0.012}        \\ \bottomrule
\end{tabular}
\vspace{-0.4cm}
\end{table}
We evaluated the advantages of using semantic segmentation to prefilter point clouds for LiDAR odometry methods. Our study focused on the KISS-ICP~\cite{kiss-icp} method, which uses iterative closest point matching methods on point clouds to correct the pose predictions of a constant velocity model. 
As this method builds on a static scene assumption, we excluded points labeled as members of a vehicle- or human-based category by different semantic segmentation models. 
In \tabref{tab:odometry}, we compare the performance of different semantic segmentation models by the absolute translational and rotational errors across all sequences in the SemanticKITTI and nuScenes validation sets. We use the default hyperparameter settings reported in~\cite{kiss-icp}.

Our results indicated that excluding points with a semantic label of a vehicle- or human-related category generally resulted in lower overall translation and rotational errors, except for the MinkowskiUNet model on the SemanticKITTI dataset. 
Among the evaluated semantic segmentation methods, filtering the point cloud based on semantic labels predicted by the PAttFormer\multitask~model produced the lowest translational and rotational errors. This was especially evident in the nuScenes datasets, which have a high density of dynamic objects per scan. We believe this is due to the model's improved accuracy on \textit{thing} class types, which was also observed in the class-wise semantic segmentation performance in \tabref{tab:nuscenes_val_semseg_per_class}.

\subsection{Qualitative Evaluations}
\label{sec:demos}
To assess the quality of our semantic segmentation results, we compared them with the predictions of the baseline multi-task network in \figref{fig:qualitative_example}. The first row depicts the predictions of two PAttFormer models, one trained for predicting object bounding boxes and one trained to classify semantic labels. 
The second row depicts the predictions of a single PAttFormer model trained in a multi-task learning setting.

The PAttFormer model in the bottom row, trained with a multi-task setting, provides more consistent semantic labels with the predicted detection bounding boxes, highlighting the mutual benefits of the perception tasks.
Upon visual inspection, the PAttFormer model predictions in the bottom row, trained with a multi-task setting, correctly detect the bendy bus and pedestrians in the left scene and the bicyclist in the right image. 
Furthermore, multi-task learning appears to result in more consistent semantic labels inside the predicted detection bounding boxes, highlighting the mutual benefits of the perception tasks.

\section{Conclusion}
\label{sec:conclusion}

We proposed a novel point-based multi-task architecture for joint semantic segmentation and object detection in point clouds. 
Our design allows hard-parameter sharing for large parts of the model, resulting in an efficient architecture that can achieve an 11Hz frame rate on a single A100 GPU. 
The effectiveness of our architecture was evaluated on two benchmarks, KITTI and nuScenes, where it achieved an mAP of 67.2$\%$ on detection and a mIoU of 79.8$\%$ on semantic segmentation. 
We observed consistent performance improvement from multi-task learning, resulting in +2.4$\%$ in detection mAP and +1.8$\%$ in segmentation mIoU compared to the single-task training. These improvements observed for only point-based architecture are even higher than observed for related multi-task architectures that use multiple point cloud representations with fewer network parts appropriable for hard parameter-sharing. 
Our data efficiency experiments further indicate that multi-task learning is particularly beneficial for smaller training datasets, where the performance improvement compared to single-task training is even more significant.
This finding can have a broader impact on robotics applications with little available training data. Future work can evaluate whether similar improvements can be observed from training with automated labels generated by vision foundation models.







{\footnotesize
\bibliographystyle{IEEEtran.bst}
\bibliography{main}
}


\end{document}